\documentclass[11pt]{article}
\usepackage{amsmath}
\usepackage[a4paper]{geometry}
\usepackage{latexsym}
\usepackage{authblk}
\usepackage{clic2017}
\usepackage{times}
\usepackage{url}
\usepackage{latexsym}
\usepackage{multicol, blindtext}

\usepackage[utf8]{inputenc}
\usepackage{array}
\usepackage{tabularx}
\usepackage{multirow}
\usepackage{graphicx}
\usepackage{amsfonts}
\usepackage{amssymb}
\usepackage{url}
\usepackage{tcolorbox}
\usepackage{enumitem}
\usepackage{graphicx}
\usepackage{multirow, makecell}
\usepackage{lingmacros}
\usepackage{bbm}
\usepackage{booktabs}
\usepackage[T1]{fontenc}
\usepackage[outline]{contour}
\usepackage{xcolor}
\contourlength{0.2pt}
\contournumber{55}%
\newcolumntype{C}{>{\centering\arraybackslash}X}

\usepackage[caption=false]{subfig}

\usepackage[T1]{fontenc}
\usepackage[outline]{contour}
\usepackage{xcolor}
\contourlength{0.2pt}
\contournumber{55}%

\usepackage{times}
\usepackage{helvet}
\usepackage{courier}
\usepackage{graphicx}

\usepackage{fancyhdr}
\usepackage{etoolbox}
\usepackage[export]{adjustbox}

\usepackage{float}
\usepackage{amsmath}

\usepackage{color}

\usepackage{booktabs,chemformula}

\title{QMUL-SDS @ SardiStance: Leveraging Network Interactions to Boost Performance on Stance Detection using Knowledge Graphs}

\author[1,2]{\textbf{Rabab Alkhalifa}}
\author[1]{\textbf{Arkaitz Zubiaga}}
\affil[1]{Queen Mary University of London, United Kingdom}
\affil[2]{Imam Abdulrahman bin Faisal University, Saudi Arabia}

\begin{document}
\maketitle
\begin{abstract}
This paper presents our submission to the SardiStance 2020 shared task, describing the architecture used for Task A and Task B. While our submission for Task A did not exceed the baseline, retraining our model using all the training tweets, showed promising results leading to (f-avg 0.601) using bidirectional LSTM with BERT multilingual embedding for Task A. For our submission for Task B, we ranked 6th (f-avg 0.709). With further investigation, our best experimented settings increased performance from (f-avg 0.573) to (f-avg 0.733) with same architecture and parameter settings and after only incorporating social interaction features- highlighting the impact of social interaction on the model's performance. 
\end{abstract}
\footnotetext{Copyright \textcopyright\ 2020 for this paper by its authors. Use permitted under Creative Commons License Attribution 4.0 International (CC BY 4.0).}


\section{Introduction}

Framed as a classification task, the stance detection consists in determining if a textual utterance expresses a supportive, opposing or neutral viewpoint with respect to a target or topic \cite{kuccuk2020stance}. Research in stance detection has largely been limited to analysis of single utterances in social media. Furthering this research, the SardiStance 2020 shared task \cite{cignarella2020sardistance} focuses on incorporating contextual knowledge around utterances, including metadata from author profiles and network interactions. The task included two subtasks, one solely focused on the textual content of social media posts for automatically determining their stance, whereas the other allowed incorporating additional features available through profiles and interactions. This paper describes and analyses our participation in the SardiStance 2020 shared task, which was held as part of the EVALITA \cite{Evalita2020} campaign and focused on detecting stance expressed in tweets associated with the Sardines movement.

\begin{figure*}[ht]
\centering
\includegraphics[width=0.95\textwidth]{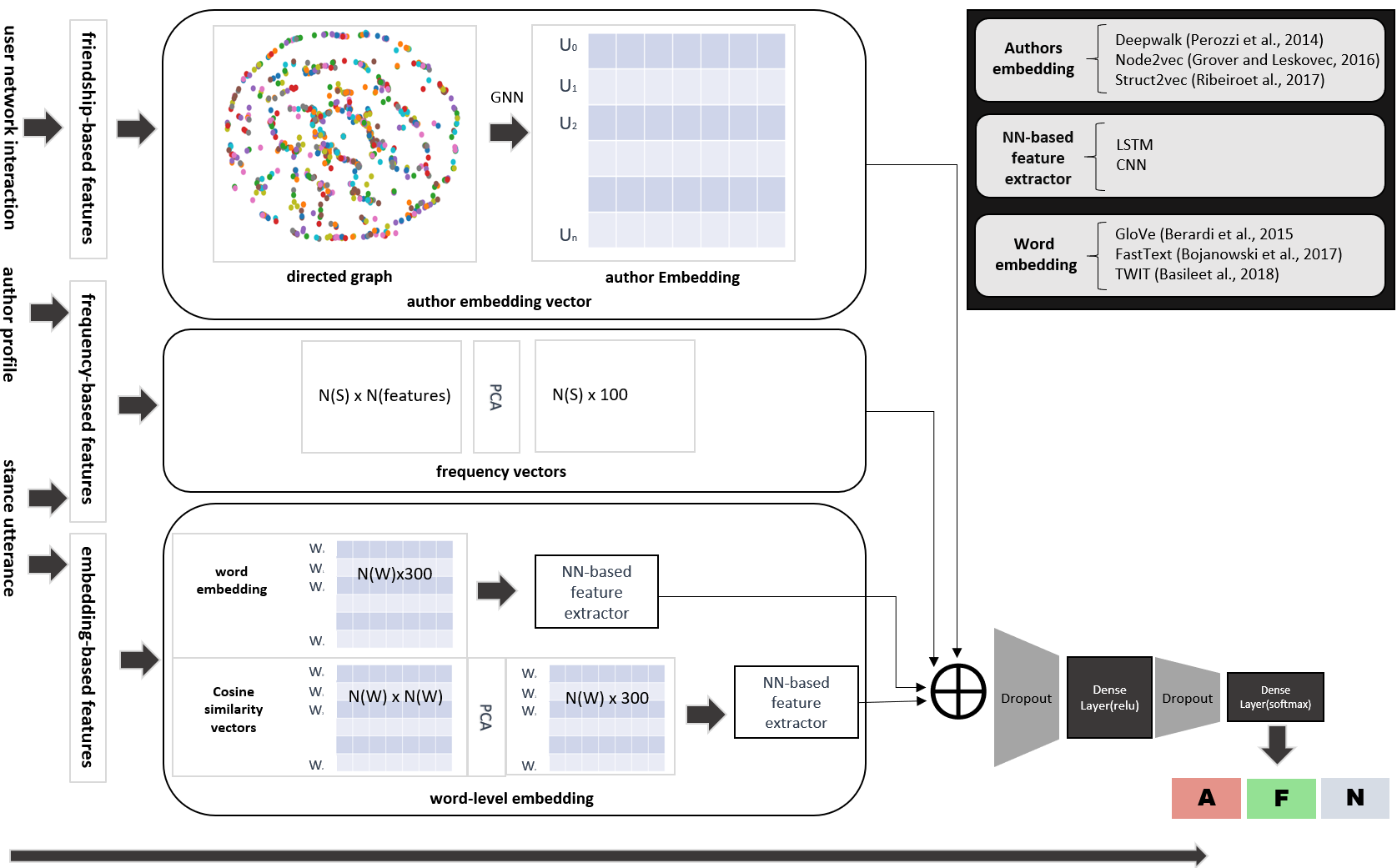}
\label{fig:general-arch}
\caption{Our framework for investigating different combinations of features.  For a network interaction graph, we generate user embeddings, using variations of graph neural network (GNN) embedding methods, namely deep-walk, struct2node and node2vec, and then concatenate author's vector with its corresponding utterance features for each stance. We also extract two types of text embedding representations for each utterance, embedding-based features, namely word embedding vectors and cosine similarity vectors, using different models including variations of CNN and bidirectional models. Further, the results of these two feature extraction methods are concatenated for the final classification step. We also consider the standard methods that extract frequency-based representations from author profiles and stance utterances including unigrams and Tfidf vectors. All these four features where combined and fed into the drop out and dense layers, to finally generate the final label using a softmax activation function. Though, we deactivate some of these four sources of features and alter the frequency-based vector by excluding some features, changing the embedding source and reducing the dimensionality for highly dimensional vectors (e.g. frequency-based features and cosine similarity vectors) using PCA.}
\end{figure*}


\section{Related Work}


In social media, \textbf{classical features} can be extracted by using \textit{stylistic signals} from text such as bag of n-grams, char-grams, part-of-speech labels, and lemmas \cite{Sobhani2019a}, \textit{structural signals} such as hashtags, mentions, uppercase characters, punctuation marks, and the length of the tweet \cite{Wojatzki2018b,Sun2016}, and \textit{pragmatic signals} related to author's profile \cite{Graells2020}. With modern deep learning models, there is shift towards \textbf{contextualised representations} using word vector representation algorithms, either by having personalised language models trained on task specific language or as a pre-trained language model offered after training using complex architecture and billions of documents. Using \textbf{deep learning layers} as automated feature engineering methods can be implemented to train the model afterwards. In \cite{augenstein2016stance}, they  utilized Bidirectional Conditional Encoding using LSTM achieving state-of-the-art results on stance detection task. Recently, there is a resurgence of research in incorporating \textbf{network homophily} \cite{Lai2017a} to represent social interactions within a network. Moreover, \textbf{Knowledge graphs} \cite{OKE2019Xu} can in turn represent these complex network relationships (e.g. authors friendships) as simple embedded vectors sampled considering the nodes and weighted edges within the network complexity structure.


\section{Definition of the Tasks}

The stance detection task has been defined in previous work as consisting in determining the viewpoint of an utterance with respect to a target topic \cite{kuccuk2020stance}, while others define it as that consisting in determining an author's viewpoint with respect to the veracity of a rumour, usually referred to as rumour stance classification \cite{zubiaga2018discourse}. SardiStance focuses on the former, and is split into two subtasks: Textual Stance Detection (Task A) and Contextual Stance Detection (Task B) \cite{cignarella2020sardistance}. Baselines are provided for Task A using SVM+unigrams as (f-avg. $0.578$), and for Task B as (f-avg. $0.628$) \cite{LAI2020101075}.

\section{Experimental Settings}\label{sec:ExpSet}


\noindent\textbf{\underline{Frequency-based features:}} \noindent These represent frequency vectors including unigram, punctuation and hashtags provided by \cite{cignarella2020sardistance}. Further, we include TFiDF vectors.

\noindent\textbf{\underline{Embedding-based features:}} \noindent \textit{\textbf{word embedding}} Italian Wikipedia Embedding \cite{berardi2015word} trained using GloVe \footnote{\url{https://github.com/MartinoMensio/it_vectors_wiki_spacy}}, Fasttext with \cite{bojanowski2017enriching} \footnote{\url{https://fasttext.cc/docs/en/pretrained-vectors.html}} trained using skip-gram model and with 300 dimensions, and TWITA embedding \cite{TWITA_in_2019embedding}. For TWITA, two versions of the same tweets were generated. One preprocessing words where each vector has 100 dimensions, provided by \cite{cignarella2020sardistance}\footnote{\url{https://github.com/mirkolai/evalita-sardistance/}} and referred to as TWITA100. The other one trained by us without any preprocessing and each vector has 300 dimensions, referred to as TWITA300. We also experimented with multilingual BERT in Task A \footnote{\url{https://tfhub.dev/tensorflow/bert_multi_cased_L-12_H-768_A-12/2}} \cite{devlin2019bert}.


\par \noindent \textit{\textbf{Cosine similarity vectors}} which was introduced previously in \cite{eger2016linearity} to encode the word meaning within the embedding space. In our work, we used TWITA300 to train the similarity vectors of all the words in the training set. 

\noindent\textbf{\underline{Network-based features:}} \noindent \textit{\textbf{Encoding users graph.}} To represent user interactions as nodes and edges, we used a counting scalar value and added one if each of the following relationships exists: friendships, retweets, quotes and replies, e.g. if all of them exist then the edge weight between two accounts is four. We calculated all the accounts provided and generate a directed complex graph conditioned by the existence of friendship, resulting in 669,745 nodes, 2,871,791 edges with an average in-degree of 4.2879 and average out-degree of 4.2879.

\par \noindent \textit{\textbf{Generating GNN Embeddings.}} Taking as input the encoded network relationships, GNN embeddings use different sampling techniques to represent every node as a vector. To extract these vectors, we experiment with different graph neural network models, namely struct2vec \cite{ribeiro2017struc2vec}, deepwalk \cite{perozzi2014deepwalk} and node2vec \cite{grover2016node2vec}.

\noindent \textbf{\underline{NeuralNetwork-based features}}
As illustrated in Figure \ref{fig:general-arch}, we have different deep learning models to extract features separately for both word embedding and similarity vectors matrices. In our work, we experiment with Convectional Neural Network (CNN) models and Long short-term memory (LSTM) models. Variations of \textbf{CNN} models where applied to NLP downstream tasks as feature extraction methods for text classification. In our work, we used two variations of CNN. In one model, we used a CNN as a one-head \textit{1D-CNN} with kernel size of 5 allowing the model to extract features with 5-grams vectors using 32 filters. Followed by a max pooling layer with pool size of 2 then flattened layer. In another model, we used a CNN as a multi-headed \textit{2D-CNN} with 1, 2, 3, 5 grams filter sizes, initialising the kernel weights with a Rectified Linear Unit (ReLU) activation function and normal distribution weights. Followed by a max pooling layer with different pooling sizes taken as one columns pooling filter with the maximum text length excluding few grams sizes. For the \textbf{LSTM}, we used two variants. One is a simple \textit{bidirectional LSTM} of 64 units followed by concatenations of max pooling and average pooling layers, and \textit{attention bidirectional LSTM} proposed by \cite{yang2016hierarchical} using 64 units followed by 128 units then attention layers\footnote{\url{https://www.kaggle.com/mlwhiz/attention-pytorch-and-keras}}.

\par \noindent \textit{\textbf{Feature Reduction.}} We experiment with different reduction length: 50, 100 and 150. Then. we set our PCA reduction to 100 as it showed best performance on evolution set. 

\par \noindent \textit{\textbf{Sentence Cleaning.}} We set the cleaning function to match the preprocessing function by \cite{cignarella2020sardistance} to generate TWITA100.


We used four final layers to receive the features and concatenate them (see Figure \ref{fig:general-arch}). In all of the experiments, our dropout layer set to $0.2$, followed by a dense layer with rule activation function and another dropout layer of $0.2$. Finally, a probability vector of the three classes is generated. To determine the correct class, we choose the one class with the highest probability. 

\section{Results}
In this section, we discuss the results of our systems submitted to the two tasks. 

For Task A, we used attention Bidirectional LSTM model performance compared to using different word embedding models, also we analysed impact of the preprocessing of the runs. Since there are too many parameters to compare with, we compared the performance of the embedding models. Our submitted models, BERT and TWITA300 illustrated in Table \ref{tbl:task1} with $*$ showed most promising results using different settings. With only \%80 training data, similarity vectors generalised better than all other embedding models. While, when all data are trained, the best model is the multilingual BERT embedding with no pre-processing (f-avg 0.601), followed by similarity vectors using cleaned text (f-avg 589).

\begin{table}[ht]
\small
\begin{center}
\begin{tabular}{l|ll|l|l}
\hline
    \multicolumn{5}{c}{\textbf{Task A} }          \\
\hline
       & \textbf{Eval. }      &        & \multicolumn{2}{c}{\textbf{Tst. f-avg} }          \\
\hline
\multicolumn{5}{c}{\textbf{Not-preprocessed} }  \\
\hline
\textbf{Emd\#}             & \textbf{\%}         & \textbf{f-avg}  & \textit{T\%80} & \textit{T\%100} \\
\hline
\textbf{\textit{BERT$^*$ }}                 & 0.480           & \textbf{0.532} & 0.533$^*$     & \textbf{0.601}   \\
\textbf{\textit{SVs}}    & \textbf{0.518}           & \textbf{0.548} & \textbf{0.589}     & 0.532   \\
\textbf{\textit{TWITA300  }}            & \textbf{0.482}           & 0.526 & \textbf{0.578}     & 0.551   \\
\textbf{\textit{TWITA100}} & 0.480           & 0.521 & 0.494     & 0.551   \\
\textbf{\textit{Fasttext}}              & 0.485           & 0.521 & 0.479     & 0.482   \\
\textbf{\textit{GloVe}}                   & 0.445           & 0.308 & 0.401     & 0.401   \\  \hline
\hline
\multicolumn{5}{c}{\textbf{Preprocessed} }  \\
\hline
\textbf{\textit{SVs}}    & \textbf{0.515}           & \textbf{0.556} & 0.524     & 0.566   \\
\textbf{\textit{TWITA100}} & \textbf{0.513}           & \textbf{0.543} & 0.560$^*$     & 0.566   \\
\textbf{\textit{FastText}}              & 0.485           & 0.489 & 0.532     & 0.528   \\
\textbf{\textit{TWITA300}}              & 0.447           & 0.490 & 0.541     & 0.506   \\
\textbf{\textit{GloVe}}                   & 0.445           & 0.308 & 0.401     & 0.401  \\
\textbf{\textit{BERT }}                 & 0.475           & 0.445 & 0.512     & 0.213   \\
\hline 
\textit{Baseline}                   &            &  & \textit{0.578}     & \textit{0.578}   \\ \hline
\end{tabular}
\end{center}
\caption{\label{tbl:task1} Results for Task A. We evaluate all the embeddings using Attention Bidirectional LSTM. Our submissions are the ones represented with $^*$. \textit{Bold fonts show results above baseline}}
\end{table}


\begin{table*}[ht]
\small
\centering
\begin{adjustbox}{max width=\textwidth}
\begin{tabular}{ l |l l |l |l |m{8cm} }
\hline
    \multicolumn{6}{c}{\textbf{Task B} }          \\
\hline
   & \multicolumn{2}{c}{\textbf{Eval.}  }   & \multicolumn{2}{|c|}{\textbf{Tst. f-avg} } &       \\ \hline
\textbf{\#M} & \%    & \textbf{f-avg} & \textit{T\%80} & \textit{T\%100} & \textbf{Settings.}       \\ \hline
1   & 0.590 & 0.651 & \textbf{0.683} & \textbf{0.733}                 & Conv2D(FastText) + Conv2D(PCA(SVs)) + PCA(unigram + Tfidf\_unigram + length) + DeepWalk \\
2   & 0.511 & 0.521 & 0.605 & 0.573                  & Conv2D( FastText ) + Conv2D( PCA(SVs) ) + PCA(unigram + Tfidf\_unigram + length) \\
\hline
3   & 0.595 & 0.640 & \textbf{0.662} & \textbf{0.719 }                 & Conv2D(FastText)+ Conv2D(PCA(SVs)) + Conv2D(PCA(Tfidf\_unigram + chargrams)) + DeepWalk  \\
4   & 0.525 & 0.507 & 0.608 & 0.604                  & Conv2D(FastText)+Conv2D(PCA(SVs))+PCA(Tfidf\_unigram + chargrams)  \\ 
\hline
5   & 0.600 & 0.645 & \textbf{0.710} & \textbf{0.718}                  & Conv2D(FastText) + Conv2D( PCA(SVs)) + PCA(unigram + length)+ DeepWalk                 \\
6   & 0.487 & 0.495 & \textbf{0.661} & 0.600                 & Conv2D(FastText + Conv2D(PCA(SVs)) + PCA(unigram + length)  \\ \hline 

7	& \textbf{0.600}	& \textbf{0.671}	& \textbf{0.709}$^*$	& \textbf{0.696}	&  Conv2D(TWITA300) + Conv2D(PCA(SVs)) + PCA( length + network\_quote\_community + network\_reply\_community + network\_retweet\_community + network\_friend\_community + userinfobio + tweetinfocreateat) + DeepWalk  \\
9	& 0.574	& 0.532	& \textbf{0.629}	& 0.615	&  Conv2D(TWITA300) + Conv2D(PCA(SVs)) + PCA( length + network\_quote\_community + network\_reply\_community + network\_retweet\_community + network\_friend\_community + userinfobio + tweetinfocreateat) \\ \hline

9 & \textbf{0.602} & \textbf{0.691} & \textbf{0.677}$^*$ & \textbf{0.681} & AttLSTM(FastText) + AttLSTM(PCA(SVs)) + PCA(puntuactionmarks + length + network\_quote\_community + network\_retweet\_community + network\_friend\_community + userinfobio) + Node2Vec \\

10 &	0.459  &	0.488  &	0.456  &	\textbf{0.660} &	AttLSTM(FastText) + AttLSTM(PCA(SVs)) + PCA(puntuactionmarks + length + network\_quote\_community + network\_retweet\_community + network\_friend\_community + userinfobio) \\ \hline

\textit{Baseline}                   &            &  & \textit{0.628}     & \textit{0.628}   \\ \hline

\end{tabular}
\end{adjustbox}
\caption{Top performing settings over all sampled runs using our architecture for Task B. Our submissions are the ones represented with $^*$. \textit{Bold fonts show highest/above baseline results}}
\label{tab:task2}
\end{table*}

For Task B, we used different feature extraction, frequency vectors, word embedding and social interaction embedding models, and monitor their performance while activating the pre-processing step in all experiments. With a diverse range of parameters, we experimented with a total of 3845 random runs. Then, we selected the best models considering macro f-score for the two classes under consideration (AGAINST and FAVOR) (f-avg). Results are shown in Table \ref{tab:task2}. By comparing our runs by adding social interaction features, our models with different settings showed a clear improvement on our models. In 1\#M, we utilise Conv2D (see NeuralNetwork-based features) for embedding vectors with  TfiDF unigram and tweet length, where the model achieved an increase on performance of (f-avg 0.16) when social interaction vectors incorporated into the model. All other models showed the same improvement  with an increase of (f-avg 0.115, 0.118, 0.081, 0.021) for 3\#M, 5\#M, 7\#M and 9\#M, respectively.

\section{Discussion and main findings}
The pipeline depicted in Figure \ref{fig:general-arch} was designed to investigate the impact of multiple features on stance detection using variations of feature extraction methods, which have been experimented in previous work but we adapted them to the Italian language in our settings. The training set contains 2132 instances with no evaluation set. In our work, we create a stratified split of 80-20 to evaluate the model, which leads to a training data with 1705 samples. Further, our investigation attempted to randomise different settings, with the aim of submitting the top two with highest f-avg score on the remaining set (Eval. 426) for both tasks. Consequently, we found that this methodology did not generalise well with the testing results. However, our main findings remain consistent across different settings when compared with our results using the stratified split (T\%80) and when the model was retrained using all the data (T\%100). While our submission evaluated both tasks separately, we discuss all conclusions jointly in this section.

Having different random settings over all frequency-based features (14, in our case) would be a bad strategy to evaluate the methods and come up with the best approach. To verify if we need to include all of these, we run an experiment by including only one feature from (unigram, Tfidf\_unigram, chargrams, network\_reply\_community, userinfobio). The selection of these features where based on selecting the best runs using only one feature from our randomised parameters. Using all the training set and CONV2D with (fasttext$,$TWEC300) and reduced SVs with deepwalk user's social interaction vector, (userinfobio$,$chargrams) achieved (f-avg 0.703 and 0.704), respectively. This is also higher than using AttLSTM for the same settings which achieved (f-avg 0.638 and 0.610). In general, we achieve better performance with CONV2D than AttnLSTM for the same settings on the test data. In another experiment, we reduced all the 14 frequency-based parameters achieving (f-avg 0.714) which performs worse than our best 3\#M (see \ref{tab:task2}). Our main conclusion is that the number of features available is not necessarily correlated with the model's performance boost.

In another experiment, we attempted to compare the performance of TWEC100 with TWEC300 (see Section \ref{sec:ExpSet}). From Table \ref{tbl:task1}, we observed that lower dimensionality and pre-processing may cause the model to under perform by around (f-avg 0.050), at least. Though, this impact was not significant with T\%100. However, matching the processing between the embedding vocabulary and the annotated set yields better performance. For example, TWITA100 was more persistent on performance between T\%80 and T\%100. This highlights the importance of pre-processing and reducing the differences between the embedding vocabularies and labelled sentences. In general, our embedding experiment for Task A show high sensitivity on model performance with pre-processing settings.

Inspired by previous work on encoding word meanings, we experimented with SVs embedding. Interestingly, these vectors showed high f-avg, better than BERT and TWITA300 with T\%80 although it showed a significant drop when the model was trained with T\%100. This finding opens an investigation towards the ability of SVs to perform better under different settings. For that, we removed PCA(SVs) and run same settings of \#M1, and our model achieved (f-avg 0.678), showing a significant impact of SVs on model's performance. Further, we investigate the robustness of deepwalk modelling over node2vec and struct2vec for the same best settings of \#M1, resulting on (f-avg 0.641 and 0.604) for node2vec and struct2vec, respectively. Also, in terms of accuracy, the deepwalk model produces an improved accuracy of (\% 0.725) compared to node2vec (\% 0.665) and struct2vec (\% 0.658). This indicates that deepwalk is more reliable on this testing set than other models.

\section{Conclusion}
In this work, we described a state-of-the-art stance detection system leveraging different features including author profiling, word meaning context and social interactions. Using different random runs, our best model achieved (f-avg 0.733) leveraging deepwalk-based knowledge graphs embeddings, FastText and similarity feature vectors extracted by two multi-headed convolutional neural networks from auther's utterance. This motivates our future, aiming to reduce the model complexity and automate the feature selection process. 

\section{Acknowledgments}
This research utilised Queen Mary's Apocrita HPC facility, supported by QMUL Research-IT.

\bibliographystyle{acl}
\bibliography{library}

\end{document}